\documentclass[letterpaper]{article} 
\usepackage{aaai25}  
\usepackage{times}  
\usepackage{helvet}  
\usepackage{courier}  
\usepackage[hyphens]{url}  
\usepackage{graphicx} 
\usepackage{natbib}  
\usepackage{caption} 
\usepackage{algorithm}
\usepackage{algorithmic}
\usepackage{booktabs} 
\usepackage{multirow}
\usepackage{amsmath}
\usepackage{amsfonts}
\usepackage{adjustbox} %
\usepackage{xcolor}         
\definecolor{plant}{RGB}{0,64,0}
\definecolor{sheep}{RGB}{128,64,0}
\definecolor{sofa}{RGB}{0,192,0}
\definecolor{train}{RGB}{128,192,0}
\definecolor{tv}{RGB}{0,64,128}
\definecolor{fan}{RGB}{128,96,128}
\definecolor{hood}{RGB}{0,160,128}
\definecolor{oven}{RGB}{160,192,192}
\definecolor{traffic}{RGB}{192,32,0}
\definecolor{sconce}{RGB}{128,160,128}
\definecolor{plate}{RGB}{192,160,128}
\definecolor{tray}{RGB}{64,160,0}
\definecolor{glass}{RGB}{0,96,128}
\usepackage{newfloat}
\usepackage{listings}
\DeclareCaptionStyle{ruled}{labelfont=normalfont,labelsep=colon,strut=off} 
\floatstyle{ruled}
\newfloat{listing}{tb}{lst}{}
\floatname{listing}{Listing}
\title{Adaptive Prototype Replay for Class Incremental
Semantic Segmentation}
\author{
    Guilin Zhu\textsuperscript{\rm 1}, Dongyue Wu\textsuperscript{\rm 1}, Changxin Gao\textsuperscript{\rm 1}\thanks{Corresponding author.}, 
    Runmin Wang\textsuperscript{\rm 2}, Weidong Yang\textsuperscript{\rm 1}, Nong Sang\textsuperscript{\rm 1}\\
}
\affiliations{
    \textsuperscript{\rm 1}School of Artificial Intelligence
 and Automation, Huazhong University of Science and Technology\\
 \textsuperscript{\rm 2}College of Information Science and Engineering, Hunan Normal University\\
 \{gzhu, dongyue\_wu, cgao, Yangwd, nsang\}@hust.edu.cn, runminwang@hunnu.edu.cn\\ 

}

\usepackage{bibentry}

\begin{document}
\maketitle

\begin{abstract}
Class incremental semantic segmentation (CISS) aims to segment new classes during continual steps while preventing the forgetting of old knowledge. Existing methods alleviate catastrophic forgetting by replaying distributions of previously learned classes using stored prototypes or features. However, they overlook a critical issue: in CISS, the representation of class knowledge is updated continuously through incremental learning, whereas prototype replay methods maintain fixed prototypes. This mismatch between updated representation and fixed prototypes limits the effectiveness of the prototype replay strategy. To address this issue, we propose the \textbf{Ada}ptive \textbf{p}ro\textbf{t}otyp\textbf{e} \textbf{r}eplay (\textbf{Adapter}) for CISS in this paper. Adapter comprises an adaptive deviation compensation (ADC) strategy and an uncertainty-aware constraint (UAC) loss. Specifically, the ADC strategy dynamically updates the stored prototypes based on the estimated representation shift distance to match the updated representation of old class. The UAC loss reduces prediction uncertainty, aggregating discriminative features to aid in generating compact prototypes. Additionally, we introduce a compensation-based prototype similarity discriminative (CPD) loss to ensure adequate differentiation between similar prototypes, thereby enhancing the efficiency of the adaptive prototype replay strategy. Extensive experiments on Pascal VOC and ADE20K datasets demonstrate that Adapter achieves state-of-the-art results and proves effective across various CISS tasks, particularly in challenging multi-step scenarios. The code and model is available at \url{https://github.com/zhu-gl-ux/Adapter}.
\end{abstract}

%

\section{Introduction}

\label{intro}
The class incremental semantic segmentation (CISS) models learn all classes through multiple steps. During each step, models focus on different classes with corresponding labels. When completing the all steps, the trained models are expected to grasp not only with the original classes but also with the newly introduced ones.  A common challenge observed in other incremental visual tasks~\cite{icarl,indetector} is \textit{catastrophic forgetting}~\cite{catastrophic}. This occurs when the model learns a new task, the network weights are fine-tuned to accommodate the new data while the old shared weights are disrupted. 

Intuitive methods~\cite{ssul,recall} attempted to alleviate catastrophic forgetting by replaying a few past images, but this suffer from storage burdens and privacy concerns. Previous CISS methods~\cite{ilt,ucd} utilized knowledge distillation (KD) to reduce catastrophic forgetting. However, these methods overlook the fact that the proportion of pixels belonging to new classes is higher than that of previously learned classes during the incremental steps, which leads to bias in the classifiers. Recent work~\cite{star} properly addressed this problem by replaying the distributions of old-class in the new-class classifiers, based on the efficient storage of representative prototypes and necessary statistics. However, the distributions replayed by static prototypes only represent the representation of previous models, neglecting the representation deviation between incremental models due to the updated shared weights.

\begin{figure}[t!]
\centering
\includegraphics[width =0.45\textwidth]{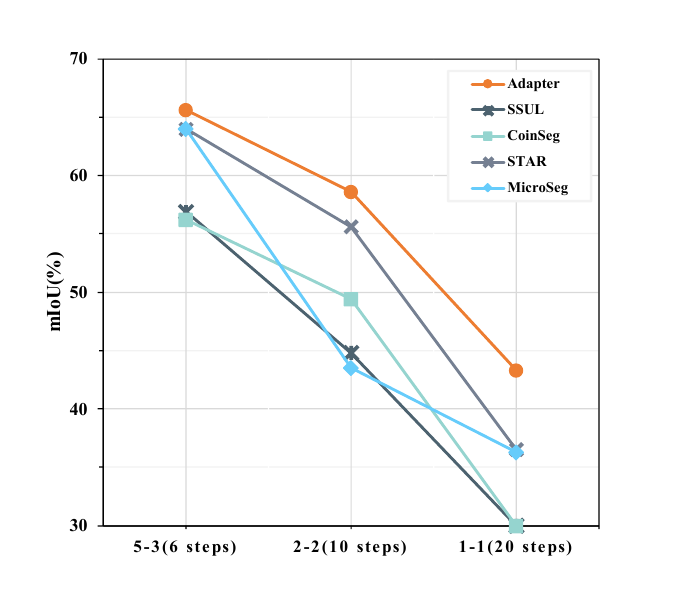}
\caption{Performance comparison of prior
works on the Pascal VOC 2012 challenging multi-step \textit{overlapped} scenarios, where contains the small classes in the initial step and long-term incremental steps. }
\label{fig1}
\end{figure}

In this paper, we propose a method called Adapter (\textbf{Ada}ptive \textbf{p}ro\textbf{t}otyp\textbf{e} \textbf{r}eplay) to tackle the issue of representation deviation in prototype replay based CISS methods. In the incremental scenario, as new classes are acquired, the models may inevitably generate representation deviations towards learned classes due to constantly updated shared weights. Therefore, Adapter avoids replaying the old-class feature distributions with fixed old-class prototypes in all the steps. Instead, Adapter exploits an adaptive deviation compensation (ADC) strategy to dynamically update representative prototypes of old classes based on the estimated representation shift distance between previous and current models in a training-free manner. Moreover, we propose an uncertainty-aware constraint (UAC) loss, which compresses the representation of each class to a compact space, thereby promoting the model to generate compact class prototypes. Additionally, to facilitate the prototype replay strategy to be carried out effectively in incremental steps, it is necessary to ensure that the representation of each class is distinguishable. Therefore, we introduce a compensation-based prototype similarity discriminative (CPD) loss. During the training process, the CPD loss repels the distance between compensatory old-class prototypes and batch-level new-class feature centers, and maintains the difference among positive new-class feature centers and negative ones. 

Our main contributions are summarized as follows: 1) We propose a method called Adapter for CISS, which adopts an adaptive deviation compensation (ADC) strategy. The ADC strategy dynamically updates the stored prototypes, thereby eliminating the representation deviation between incremental models. 2) We design an uncertainty-aware constraint (UAC) loss to decrease the uncertainty predictions and aggregate class representations, facilitating the generation of compact prototypes. 3) We introduce a compensation-based prototype similarity discriminative (CPD) loss to enhance the discrimination of each class prototype and improve the efficiency of the adaptive prototype replay strategy. 4) We demonstrate the effectiveness of our Adapter through extensive experiments on two public CISS benchmarks and achieve state-of-the-art performance, particularly in more challenging and realistic multi-step  scenarios (Fig.~\ref{fig1}).

\section{Related work}
\label{rw}
\subsection{Continual learning}
Continuous learning (CL), also referred to as incremental learning~\cite{increlearning}, seeks to acquire new knowledge from continuous data streams while minimizing the risk of forgetting previously learned concepts. In recent years, research on continuous learning has concentrated on image classification, which can be broadly categorized into regularized, architectural, and replay methods. Regularization methods utilize regularization or distillation techniques~\cite{kd} to address catastrophic forgetting at both the parameter~\cite{ewc,si} and function levels~\cite{lwf,podnet}. Architectural methods~\cite{AGS,den} involve selecting isolated parameter subspace for different tasks. Replay methods~\cite{exam,replay} has focused on enhancing replay effectiveness. For instance,~\cite{meic} utilizes low-fidelity auxiliary samples of old-class knowledge transfer to enhance memory efficiency.~\cite{sdc} estimates feature drift caused by learning new tasks and compensates for the prototypes of previous tasks using an embedding network to enhance performance in image classification. Our method differs from prior approaches in two key aspects. First, we update the prototypes used for replay incrementally by estimating and compensating for representation deviation between incremental models without additional training. Second, while prior methods generally address task-agnostic incremental image classification, our approach specifically targets model representation deviation in class-incremental semantic segmentation with uncertain pseudo-predictions.

\subsection{Class incremental semantic segmentation}
ILT~\cite{ilt} extends incremental learning to semantic segmentation by implementing distillation techniques to the output layer and intermediate features. MiB~\cite{mib} addresses the unique issue in CISS called \textit{background shift} by modeling the background semantic during learning steps. PLOP~\cite{plop} proposes a multi-scale local distillation scheme to alleviate catastrophic forgetting. SSUL~\cite{ssul} and MicroSeg~\cite{microseg} define unknown classes within the background class to enhance plasticity, while replacing cross-entropy (CE) loss with binary cross-entropy (BCE) loss to mitigate catastrophic forgetting. DKD~\cite{dkd} proposes the decoupled knowledge distillation to improve the performance. IDEC~\cite{Idec} and CoinSeg~\cite{coinseg} utilize contrast learning to obtain discriminative representation. In addition, several methods~\cite{alife,star} explored the efficiency of no-exemplar memory replay. For instance, STAR~\cite{star} stored a compact prototype of each class and the necessary statistics used to replay the feature distributions in subsequent steps, addressing classifier bias. However, failing to account for the representation deviation generated by the incremental models on old-class knowledge limits the effectiveness of prototype replay. In contrast, our method replays old-class distributions based on considering the representation deviation of incremental models, thereby providing a distinct advantage.
\begin{figure*}[t]
{\includegraphics[width =1\textwidth]{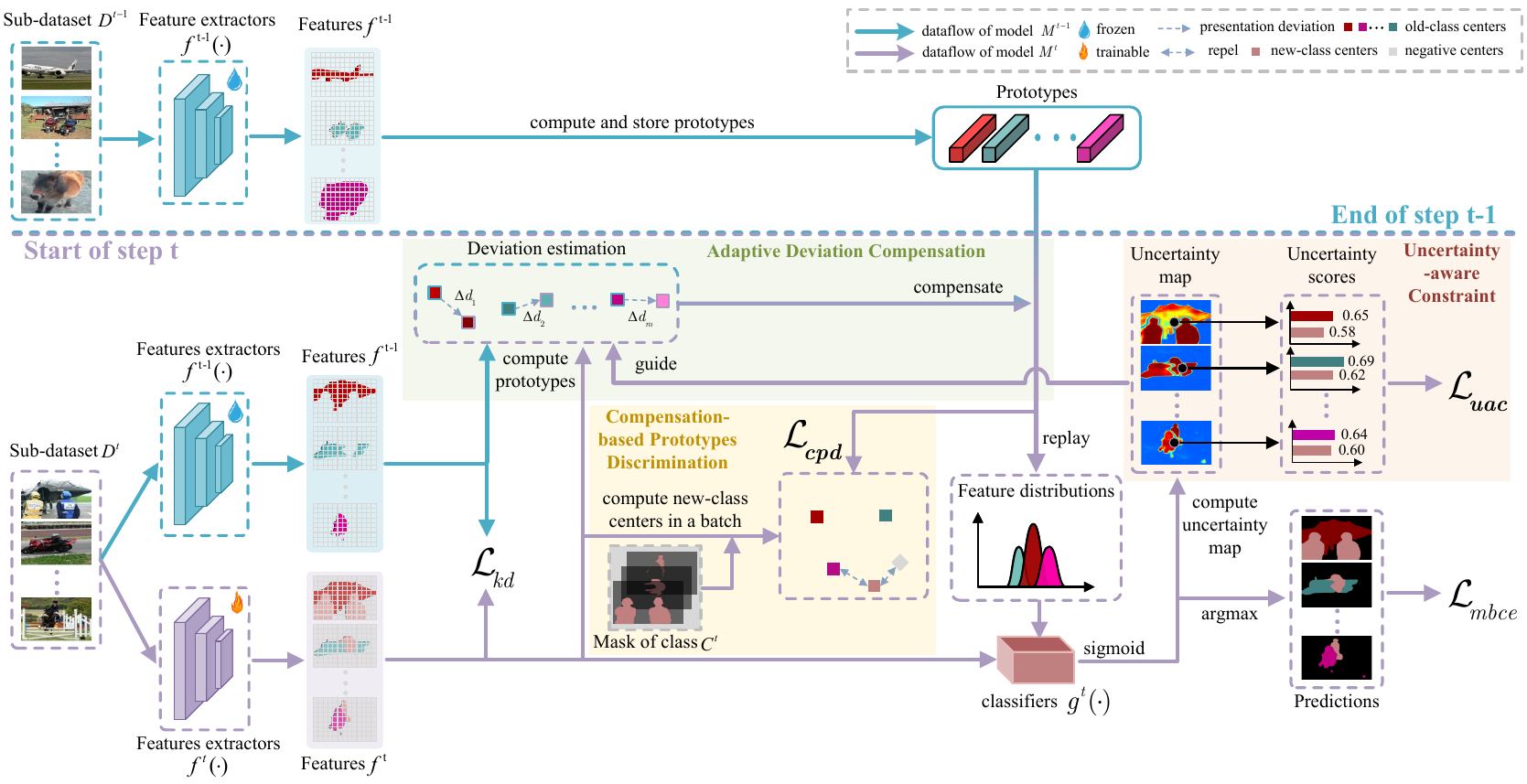}}
\caption{Overview of the proposed Adapter. After training of step $t-1$, old-class prototypes 
 are computed and saved. At the current step $t$, representation deviations towards old-class are estimated and prototypes are updated with the ADC strategy. Old-class feature distributions are replayed with updated prototypes for classifiers $g^t\left(\cdot\right)$. The UAC loss enhances the consistency of the same class by aggregating the representation of each class with uncertainty-aware loss. The CPD loss facilitates discrimination between new-class features and updated old-class prototypes, improving the efficiency of prototype replay.}
\label{fig2}
\end{figure*}

\section{Method}
\label{method}
\subsection{Problem definition}
In CISS, the model training process involves a series of incremental steps, as $t=1,\ldots,T$. At each incremental step $t$, a sub-dataset $D^{t}$ is used to learn the classes $C^{t}$ (full classes as $C$) for the current step. The sub-dataset $D^{t}$ consists of $N^{t}$ pairs set $\{x_{i}^{t},y_{i}^{t},\}_{i=1}^{N^{t}}$, where $x_i^t\in\mathbb{R}^{H\times W\times3}$ denotes the image with size of $H\times W$ and $y_i^t\in\mathbb{R}^{H\times W}$ denotes the corresponding ground-truth mask. It is worth noting that the labels $\{y_i^t\}_{i=1}^{N^t}$ within sub-dataset $D^{t}$ only consider the current learning classes pixels as foreground classes $C^{t}$, while other classes are treated as background. We define the model in step $t$ as $M^{t}$, which consists of feature extractors $f^t\left(\cdot\right)$ and classifiers $g^t\left(\cdot\right)$. For an input image $\mathbf{x}_i^t$, the output of model is $\hat{\mathbf{y}}_{i}^t\in\mathbb{R}^{H\times W}$ that computed by $\hat{\mathbf{y}}_{i}^t=M^t\left(x_i^t\right)=g^t\circ f^t\left(\mathbf{x}_{i}^t\right)$. We denotes extracted features as $f_i^t\in\mathbb{R}^{h\times w\times d}$ computed by $f_i^t=f^t\left(\mathbf{x}_{i}^t\right)$, where $d$ represents the dimension of features. After completing all incremental steps, the model is desired to have considerable performance not only on old classes but also on new ones.

\subsection{Overview}

We show the overview of our proposed Adapter in Fig~\ref{fig2}. After the training process of step $t-1$, the model inference all training samples in sub-dataset $D^{t-1}$ to obtain and store class prototypes in $C^{t-1}$ and essential statistics. At early training epochs of steps $t$, the saved old-class prototypes are used for replaying feature distributions of old-class as additional background samples in classifier $g^t\left(\cdot\right)$. As parameters of model $M^{t}$ are updated to adapt new-class, the representation of old-class appears deviation with high probability. We introduce the ADC strategy to estimate the representation deviation of each old-class and dynamically update saved prototypes with deviation compensation. In addition, we use UAC and CPD loss to constrain prediction uncertainty while enhancing the distinguish ability between new-class features and old-class prototypes. Consequently, the integrated objective of each incremental step is defined as:
\begin{equation}\mathcal{L}=\mathcal{L}_{mbce}+\alpha\mathcal{L}_{kd}+\beta\mathcal{L}_{uac}+\gamma\mathcal{L}_{cpd},\end{equation}
where $\mathcal{L}_{mbce}$ and $\mathcal{L}_{kd}$ denote multiple binary cross-entropy (mBCE) loss and knowledge distillation loss as in~\cite{ssul, star} respectively. $\alpha$, $\beta$, $\gamma$ are  hyperparameters to balance the weights among above terms. 

\subsection{Adaptive Deviation Compensation}
After the training process of step $t-1$, we follow the prototype replay strategy~\cite{star} to store the old-class prototypes and necessary statistics for replaying the gaussian-distributed old-class features. These replayed features are treated as additional background samples to mitigate the bias in the new-class classifiers.However, the representation of old-class features exhibits deviation due to the gradient-updated parameters within different new training data distributions. This renders the fixed prototype replay strategy of limited effectiveness.

Intuitively, the deviation in representation should be computed to correct the replayed feature distributions of old classes. Unfortunately, due to ethical and privacy considerations, the old data is unavailable. As a result, we indirectly estimate the representation deviation in the current incremental step using a limited number of old-class pixels present in the current new-class training dataset (e.g., an image containing a \textit{car} (old class) and a \textit{person} (new class)). 

Specifically, after several training epochs in step $t$, we first obtain the prediction of each image with current model by:
\begin{equation}\label{eq2}
~\bar{y}_{i}^{t} =\begin{cases}\hat{y}_{i}^{t} & \mathrm{~if~~}y_{i}^{t}=c_{b}^{t}\wedge \phi_{i}^{t}\geq\tau
\\0 & \mathrm{~otherwise~}\end{cases},
\end{equation}
where $c_{b}^{t}$ represents the background class that contains old-class and truth background in step $t$, and $\wedge$ represents the co-taking of conditions. ${\phi}_{i}^{t}$ denotes the certainty scores of prediction, the details of which will be given in Eq.~\ref{eq9}. Analogously, the prediction $\bar{y}_{c}^{t-1}$ is obtained by replacing ${\phi}_{i}^{t}$ with ${\phi}_{i}^{t-1}$ in Eq.~\ref{eq2}. $\tau$ is a hyperparameter to screen out the high-confidence old-class predictions. 

Since the predictions of the current model differ from the previous predictions, computing the sub-prototype of each old class using the respective predictions results in misaligned distributions. Therefore, we generate the unified prediction masks by:
\begin{equation}\hat{y}_{i}^{C^{t-1}}=\begin{cases}\bar{y}_{i}^{t} & \mathrm{~if~}\bar{y}_{i}^{t}=\bar{y}_{i}^{t-1}
\\0 & \mathrm{~otherwise~}\end{cases}.\end{equation}

Next, we calculate the feature centers of old-class on current data space. We utilize previous feature extractors and current ones to capture features $\{f_{i}^{t-1}\}_{i=1}^{N^{t}}$ and $\{f_{i}^{t}\}_{i=1}^{N^{t}}$, where $f_{i}^{t-1}\in\mathbb{R}^{h\times w\times d}$ is computed by $f_{i}^{t-1}=f^{t-1}\left({\mathbf{x}_{i}^t}\right)$ and $f_{i}^{t}$ is so on. We perform the average operation on normalized features of previous extractors with unified prediction masks to obtain sub-prototype $\hat{P}_{c}^{t-1}\in\mathbb{R}^{d}$ of each old-class by:
\begin{equation}\label{eq4}\hat{P}_c^{t-1}=\frac{\sum_{i=1}^{N^{t-1}}\sum_{j=1}^{h\times w}(\hat{f}_{i,j}^{t-1}\odot\delta\{\tilde{y}_{i,j}^{{C}^{t-1}}=c\})}{\left\|\sum_{i=1}^{N^{t}}\sum_{j=1}^{h\times w}(\hat{f}_{i,j}^{t-1}\odot\delta\{\tilde{y}_{i,j}^{{C}^{t-1}}=c\})\right\|_2}, \end{equation}
where $j$ represents the spatial location, $\odot$ and $\delta\{\cdot\}$ indicate element-wise multiplication and indicator function respectively. $\tilde{y}_{i}^{C^{t-1}}$ and $\hat{f}_{i}^{t-1}$ denote downsampled unified masks and features. $\|\cdot\|_2$ represents L2-norm. Analogously, the sub-prototype $\hat{P}_{c}^{t}$ is obtained by replacing $\hat{f}_{i,j}^{t-1}$ with $\hat{f}_{i,j}^{t}$ in Eq.~\ref{eq4}.

With $\hat{P}_{c}^{t-1}$ and $\hat{P}_{c}^{t}$ obtained from old and new model inference on the same prediction region, we intuitively compute the deviation towards old-class in current sub-dataset by:
\begin{equation}\triangle_{c^{t-1}}^{t-1\to t}=\vec{\Omega}(\hat{P}_{c}^{t-1},\hat{P}_{c}^{t}),\end{equation}
where $\vec{\Omega}(\cdot,\cdot)$ indicates the displacement vector difference operation. Hence, the shifted prototype of each old-class is estimated by:
\begin{equation}\grave{P}_{c}^{t-1}=P_{c}^{t-1}+\triangle_{c^{t-1}}^{t-1\to t},\end{equation}
where $P_{c}^{t-1}$ is stored prototype of old-class $c$. However, there exists uncertainty in the prediction masks $\hat{y}_{i}^{C^{t-1}}$ and statistical limitation of sub-prototype compared to the certain prototype with ground truth mask and abundant statistic information in the previous steps. To ensure the robustness of replayed features, we combine the shifted prototypes and stored prototypes with adaptive weighted average to obtain the compensatory prototype of each old-class by:
\begin{equation}\bar{P}_c^{t-1}={\rho}_c^{t-1}\grave{P}_{c}^{t-1}+(1-{\rho}_c^{t-1})P_{c}^{t-1}.\end{equation}
Here, ${\rho}_c$ is an adaptive weight to consider the contribution between two terms, which is formulated by:
\begin{equation}
\rho_{c}^{t-1}=\frac{\sum_{i=1}^{N^{t}}\sum_{j=1}^{h\times w}\delta\{\tilde{y}_{i,j}^{C^{t-1}}=c\}}{\eta_{c}^{1:t-1}+\sum_{i=1}^{N^{t}}\sum_{j=1}^{h\times w}\delta\{\tilde{y}_{i,j}^{C^{t-1}}=c\}},\end{equation}
where $\eta_{c}^{1:t-1}$ denotes the sum of pixels belonging to old-class $c$ (contain the labels and unified prediction masks in previous $1~,...,~t-1$ steps). 

The compensatory prototypes of the old classes, obtained through a training-free approach, serve to generate sufficient feature distributions that are then replayed in the current new-class classifiers. These distributions not only encompass the original distributions of the old data but also account for the representation deviation between the incremental models. With these conditions in place, the adaptive deviation compensation strategy enhances the efficiency and robustness of the prototype replay.
\begin{table*}[t!]
  \centering  
\footnotesize
\begin{tabular}{c|ccc|ccc|ccc|ccc|ccc}
\hline
\multirow{2}*{Method} 
     & \multicolumn{3}{c}{15-1 (6 steps)} &  \multicolumn{3}{|c|}{5-3 (6 steps)} & \multicolumn{3}{c|}{10-1 (11 steps)} & \multicolumn{3}{c|}{2-2 (10 steps)} & \multicolumn{3}{c}{1-1 (20 steps)}   \\ 
    & old & new & all & old & new & all & old & new & all & old & new & all & old & new & all\\
    \hline

    MiB(CVPR20) & 35.1 & 13.5 & 29.7 & 57.1 & 42.6 & 46.7 & 12.3 & 13.1 & 12.7 & 41.7 & 26.0 & 28.2 & 38.5 & 8.1 & 11.0 \\

    PLOP(CVPR21) & 65.1 & 21.1 & 54.6 & 41.1 & 23.4 & 25.9 & 44.0 & 15.5 & 30.5 & 24.1 & 11.9 & 13.7 & 12.4 & 11.9 & 4.7 \\

    SSUL(NeurIPS21) & 77.3 & 36.6 & 67.6 & 72.4 & 50.7 & 56.9 & 71.3 & 46.0 & 59.3 & 61.4 & 42.1 & 44.8 & 52.6 & 27.5 & 29.9 \\

    DKD(NeurIPS22) & 78.1 & 42.7 & 69.7 & 69.6 & 53.5 & 58.1 & 73.1 & 46.5 & 60.4 & 60.5 & 45.8 & 47.9 & \underline{56.1} & 24.6 & 27.6 \\

    RCIL(CVPR22) & 70.6 & 23.7 & 59.4 & 65.3 & 41.5 & 50.3 & 55.4 & 15.1 & 34.3 & 28.3 & 19.0 &19.4 & - & - & - \\

    MicroSeg(NeurIPS22) & \underline{80.1} & 36.8 & 69.8 & \textbf{77.6} & 59.0 & 64.3 & 72.6 & 48.7 & 61.2 & 61.4 & 40.6 & 43.5  & 55.9 & 34.2 & 36.3 \\

    IDEC(PAMI23) & 77.0 & 36.5 & 67.3 & 67.1 & 49.0 & 54.1 & 70.7 & 46.3 & 59.1 & - & - & - & - & - & - \\

    CoinSeg(CVPR23) & \textbf{80.6} & 36.2 & 70.1 & 66.8 & 51.9 & 56.2 & 73.7 & 45.0 & 60.1 & \textbf{68.3} & 46.2 & 49.4 & 53.7 & 27.4 & 29.9 \\

    STAR(NeurIPS23) & 79.4 & \underline{50.3} & \underline{72.5} & 71.9 & \underline{61.5} & \underline{64.4} & \underline{73.1} & \textbf{55.4} & \underline{64.7} & 59.2 & \underline{55.0} & \underline{55.6} & 43.6 & \underline{35.7} & \underline{36.5} \\

    Ours & 79.9 & \textbf{51.9} & \textbf{73.2} & \underline{73.8} & \textbf{61.9} & \textbf{65.3} & \textbf{74.9} & \underline{54.3} & \textbf{65.1} & \underline{62.8} & \textbf{57.9} & \textbf{58.6} & \textbf{63.4} & \textbf{40.5} & \textbf{42.7} \\
\hline
\end{tabular}

\caption{
    Quantitative results on the validation set of PASCAL VOC~\cite{pascal} for \textit{overlapped} settings. The best and second best performances are in bold and underline, respectively.
  }
  \label{tab1}
\end{table*}
\subsection{Uncertainty-Aware Constraint}
\label{uncer}
As a standard practice, pseudo-labels generated from the previous model are employed to tackle the problem of background shift. Specifically, the background values in the current labels at the current step are replaced by the predictions from previous models using a threshold filtering strategy, which may be either fixed~\cite{plop} or dynamic~\cite{Idec}. However, these strategies focus solely on the maximum value of the logistic outputs, overlooking the interrelationships within the logistic distributions. This introduces uncertainty in the pseudo-labels, which hampers the effective updating of prototypes stored from previous steps.

To address this issue, we propose an uncertainty-aware constraint loss that takes into account the interrelationships among class logistic outputs. This loss encourages the model to produce compact representations for pixels belonging to the same old class. At step $t$, we first compute the certainty scores of predictions mentioned in Eq.~\ref{eq2}. 

Specifically, let $\hat{S}_i^t\in\mathbb{R}^{H,W,C^1+\cdotp\cdotp\cdotp+C^t}$ denote as the logistic outputs of image $i$ obtained by the current model (which include all classes need to be learning in current step $t$). We compute the certainty scores of predictions by:
\begin{equation}\label{eq9}
\phi_i^{t}=\Lambda(\varphi(\hat{S}_i^t),1)-\Lambda(\varphi(\hat{S}_i^t),2),
\end{equation}
where $\Lambda(\cdot)$ represents the maximum heap sorting and selection function, and $\varphi(\cdot)$ denotes the sigmoid function. Since our primary focus is on enhancing low-probability predictions, high-probability predictions need to be filtered out to avoid overconfidence.
We compute the masks by utilizing the ground truth and predictions:
\begin{equation}
m_i^t=\begin{cases}
    0~~~\mathrm{if}~(y_{i}^{t}=\hat{y}_i^t)\in C^t~\mathrm{or}~max(\varphi(\hat{S}_i^t))\geq\tau\\
     1~~~\mathrm{otherwise}
\end{cases}.
\end{equation} 

Finally, we achieve the UAC loss as:
\begin{equation}\mathcal{L}_{uac}=d(u_i^t\odot m_i^t,\Omega_i^t),\end{equation}
where $d(\cdot)$ denotes a distance measurement function, $u_i^t = 1-\phi_i^t$ is denoted as the uncertainty scores of predictions, and $\Omega_i^t$ is the ideal targets of uncertainty. With $\mathcal{L}_{uac}$, the uncertainty in predictions is mitigated, features of the same class are aggregated to enhance model rigidity.

\subsection{Compensation-based Prototype Discrimination}
The previously proposed ADC strategy and UAC loss successfully address representation deviation and prediction uncertainty during incremental steps. However, feature extractors may generate indistinguishable features in two scenarios: when background pixels resemble the foreground and when foreground pixels resemble the background. These indistinguishable features lead to confusion in current classifiers for new classes. To mitigate this issue, we propose a compensation-based prototype similarity discriminative loss to distinguish between similar representations of new and old classes.

Specifically, we consider the similarity class issue from two perspectives. First, for pixels of new classes at the current step, we can leverage ground truth labels to obtain the centers of new-class features within a batch:
\begin{equation}\zeta_{c}^t=\frac{\sum_{i\in \mathcal{B}}\sum_{j=1}^{h\times w}(\hat{f}_{i,j}^t\odot\delta\{\tilde{y}_{i,j}^t=c\})}{\left\|\sum_{i\in \mathcal{B}}\sum_{j=1}^{h\times w}(\hat{f}_{i,j}^t\odot\delta\{\tilde{y}_{i,j}^t=c\})\right\|_2},\end{equation}
where $\mathcal{B}$ denotes the size of batch. Intuitively, we expect to discriminate them with the compensatory old-class prototypes:
\begin{equation}\mathcal{L}_{n{\leftrightarrow}o}=\frac1{\left|C^t\right|}\sum_{c\in C^t}\frac{1} {\underset{c\in C^{1:t-1}}{min}\left\|\zeta^t-\bar{P}^{t-1}\right\|_2+\epsilon},\end{equation}
where $\epsilon$ is a constant to avoid zero denominator. On the other hand, for the pixels belonging to the background classes (containing old or feature classes) that are misclassified as foreground classes in the current step, we believe that these pixels are similar to the new classes, thus confusing the new-class classifiers. We compute the feature centers of these misclassified pixels:
\begin{equation}\check{\zeta}_c^t=\frac{\sum_{i\in\mathcal{B}}\sum_{j=1}^{h\times w}(\hat{f}_{i,j}^t\odot\delta\{\tilde{y}_{i,j}^t\ne c~\wedge~ \hat{y}_{i,j}^t=c\})}{\left\|\sum_{i\in\mathcal{B}}\sum_{j=1}^{h\times w}(\hat{f}_{i,j}^t\odot\delta\{\tilde{y}_{i,j}^t\ne c~\wedge~ \hat{y}_{i,j}^t=c\})\right\|_2}.\end{equation}
Accordingly, We also add a loss term to separate the features of these 
background pixels from the features of foreground pixels:
\begin{equation}\mathcal{L}_{n_{pos}\leftrightarrow n_{neg}}=\frac{1}{\left|C^{t}\right|}\sum_{c\in C^{t}}\frac{1}{\left\|\zeta_c^{t}-\check\zeta_c^{t}\right\|_{2}+\epsilon}.\end{equation}
Combining the above two loss terms, we can obtain the final CPD loss:
\begin{equation}
\mathcal{L}_{cpd}=\mathcal{L}_{n{\leftrightarrow}o}+\mathcal{L}_{n_{pos}\leftrightarrow n_{neg}}.
\end{equation}

Using $\mathcal{L}_{cpd}$, the features of new classes are not only separated from the compensatory prototypes of old classes but also from the features of old classes within the same batch. Additionally, the CPD loss motivates the model to learn discriminative representations for new classes, which plays a crucial role in storing representative prototypes for new classes at the current step.
\begin{table*}[t!]
  \centering
  \footnotesize
  \begin{tabular}{c|ccc|ccc|ccc|ccc}
    \hline
    \multirow{2}*{Method} 
     & \multicolumn{3}{c}{100-50 (2 steps)} &  \multicolumn{3}{|c|}{50-50 (3 steps)} & \multicolumn{3}{c|}{100-10 (6 steps)} & \multicolumn{3}{c}{100-5 (11 steps)}   \\ 
      & old & new & all & old & new & all & old & new & all & old & new & all\\
    \hline
    MiB(CVPR20) & 40.5 & 17.2 &32.8 & 45.6 & 21.0 & 29.3 & 38.2 & 11.1 & 29.2 & 36.0 & 5.7 & 26.0 \\

    PLOP(CVPR21) & 41.9 & 14.9 & 32.9 & 48.8 & 21.0 & 30.4 & 40.5 & 13.6 & 31.6 & 39.1 & 7.8 & 28.8 \\

    SSUL(NeurIPS21) & 41.3 & 18.0 & 33.6 & 48.4 & 20.2 & 29.6 & 40.2 & 18.8 & 33.1 & 39.9 & 17.4 & 32.5 \\

    RCIL(CVPR22) & 42.3 & 18.8 & 34.5 & 48.3 & 24.6 & 32.5 & 39.3 & 17.7 & 32.1 & 38.5 & 11.5 & 29.6 \\

    IDEC(PAMI23) & 42.0 & 18.2 & 34.1 & 47.4 & 26.0 & 33.1 & 40.3 & 17.6 & 32.7 & 39.2 & 14.6 & 31.0 \\

    STAR(NeurIPS23) & \underline{42.4} & \textbf{24.2} & \underline{36.4} & \underline{48.7} & \underline{27.2} & \underline{34.4} & \underline{42.0} & \textbf{20.6} & \underline{34.9} & \underline{41.7}   & \underline{17.5} & \underline{33.7}\\

    Ours & \textbf{43.1} & \underline{23.6} & \textbf{36.7} & \textbf{49.3} & \textbf{27.3} & \textbf{34.7} & \textbf{42.9} & \underline{19.9} & \textbf{35.3} & \textbf{42.6} & \textbf{18.0} & \textbf{34.5} 
 \\
    \hline
  \end{tabular}
  \caption{
    Quantitative results on the validation set of ADE20K~\cite{Ade} for \textit{overlapped} settings. The best and second best performances are in bold and underline, respectively.
  }  \label{tab2}  
\end{table*}

\begin{figure*}[t!]
\centering
\includegraphics[width =0.88\textwidth]{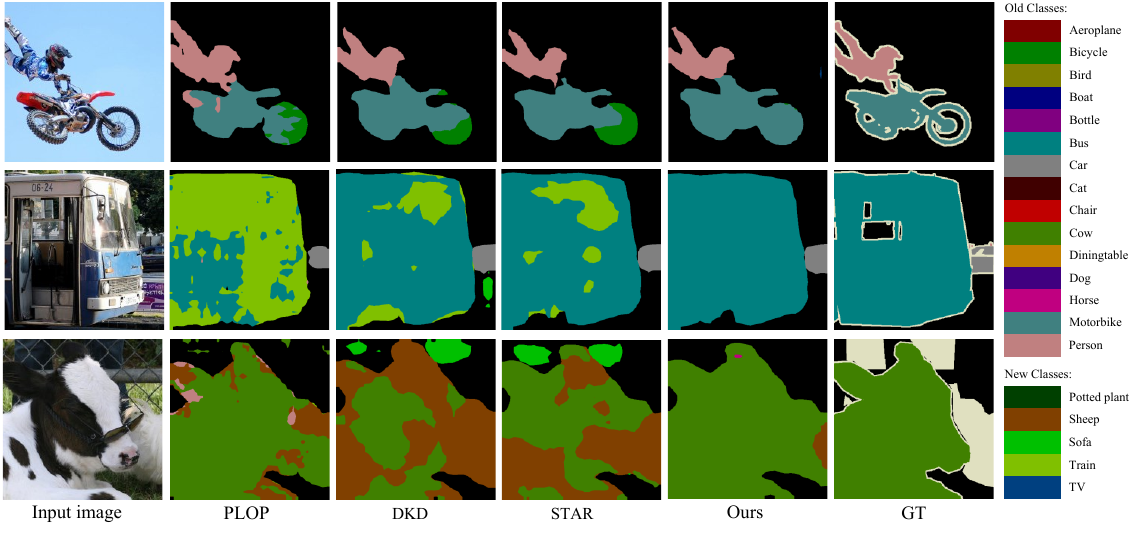}
\caption{Qualitative comparison on Pascal VOC 2012 between Adapter and previous methods.}
\label{fig3}
\end{figure*}

\section{Experiments}
\label{ex}
\subsection{Experimental Setup}
\textbf{Datasets and Evaluation Metrics.} We evaluate our proposed Adapter on Pascal VOC 2012~\cite{pascal} and ADE20K~\cite{Ade}. Pascal VOC 2012 consists of 20 foreground classes and one background class while ADE20K is a large-scale semantic segmentation dataset containing 150 classes. For evaluation, we fellow the work\cite{plop} to use mean Intersection over Union (mIoU) as the evaluation metric.

\textbf{Protocol.} Following~\cite{dkd,star}, we mainly consider the \textit{overlapped} settings in CISS, which means the background pixels in step $t$ might contain old and future classes. We denote by $(N_1-N_2)$ the incremental scenario,
where $N_1$ and $N_2$ are class numbers of the initial step and each incremental step, respectively. For each
benchmark dataset, we follow~\cite{dkd} to evaluate our method under multiple incremental scenarios (i.e., 19-1, 15-5, 15-1, 5-3, and 10-1 in Pascal VOC, and 100-50, 50-50, 100-10 in ADE20K). In addition, we also evaluate our method on more challenging incremental scenarios (i.e., 2-2 and 1-1 in Pascal VOC, and 100-5 in ADE20K).

\textbf{Implementation Details.}
We follow the common practice~\cite{mib} to use DeepLabv3~\cite{deeplab} with
ResNet-101~\cite{deep} pre-trained on ImageNet as the segmentation network. In line with~\cite{plop,alife}, we utilize different training strategies
for two datasets. For Pascal VOC 2012, we set the initial learning rate to $10^{-3}$ and $10^{-4}$ for training with 60 epochs in the initial step and incremental steps, respectively. For ADE20K, we train for 100 epochs with an initial learning rate of $2.5\times 10^{-4}$ for the initial step and $2.5\times 10^{-5}$ for incremental steps. The model is optimized by SGD with a momentum of 0.9, and the batch size is set to 24. Hyper-parameters $\alpha,\beta,\gamma, and~\tau $ are set to 5, 0.1, 0.05, and 0.7, respectively. We run our experiments on
NVIDIA RTX 4090 GPUs using PyTorch.

\subsection{Experimental Results}
\textbf{Comparison on Pascal VOC.} We evaluate our Adapter on the Pascal VOC 2012 dataset across various incremental scenarios, including standard settings in other methods (e.g., 15-1, 5-3, and 10-1) and two challenging long-term scenarios, namely 2-2 and 1-1, which involve continuously introducing new classes. The comparative results of our method against classical and prior CISS methods are presented in Tab.~\ref{tab1}. We denote "old" as the mIoU of the classes containing the background class in the initial step, and "new" as the mIoU of the classes for all incremental steps. The results demonstrate that our method consistently delivers superior performance across various incremental scenarios. Especially in the challenging multi-step incremental settings, such as the \textit{overlapped} 2-2 (10 steps) and 1-1 (steps) tasks, our method surpasses the state-of-the-art
by 3.0\% and 6.2\% in terms of mIoU, respectively.  Fig.~\ref{fig3} shows qualitative results of our approach compared
with other competitors on an \textit{overlapped} 15-1 setting, highlighting the superiority of our method.  For example, our method accurately classifies the tires of the \textit{motorbike} in row 1, unlike several competitors, who misidentify these pixels as part of bicycles. Additionally, our method distinctly differentiates the \textit{bus} (in row 2) and the \textit{cow} (in row 3) from the similar-looking \textit{train} and \textit{sheep} classes. In contrast, the \textit{train} and \textit{sheep} classes, added during incremental steps, lead to catastrophic forgetting of similar classes in competing methods.
\begin{table}[t!]
  \centering
  \footnotesize
  \begin{tabular}{c|c|c|c|ccc}
    \hline
    \multirow{2}*{Baseline} & \multirow{2}*{ADC}& \multirow{2}*{UAC}& \multirow{2}*{CPD}&\multicolumn{3}{c}{15-1 (6 steps)}\\
    & & & & old & new & all\\
    \hline
    \checkmark & & & & 78.7 & 47.4 & 71.3\\
    \checkmark & \checkmark& & & 79.2 & 48.5 & 71.9 \\
    \checkmark & \checkmark &  \checkmark&  & 79.5 & 49.8 & 72.4\\
    \checkmark & \checkmark& \checkmark & \checkmark& 79.9 & 51.9 & \textbf{73.2}\\
    \hline
  \end{tabular}
  \caption{
    Ablation study of our method components on PASCAL VOC \textit{overlapped} 15-1 (6 steps).
  }
  \label{tab3}  
\end{table}

\textbf{Comparison on ADE20K.} Table~\ref{tab2} presents the comparison results on the ADE 100-50, 50-50, 100-10, and 100-5 tasks. Despite the larger class number in ADE20K, our method once again outperforms the previous state-of-the-art. Specifically, for the ADE20K \textit{overlapped} 100-5 (11 steps) setting, we obtain a 34.5 mIoU, which is a 0.8 gain over the second-best method. This further demonstrates the capability of our method to address more challenging scenarios.
\begin{table}[t!]
  \centering
  \footnotesize
  \begin{tabular}{c|c|c|ccc}
    \hline
      \multirow{2}*{Copro} & \multirow{2}*{$\mathcal{L}_{n{\leftrightarrow}o}$} & \multirow{2}*{$\mathcal{L}_{n_{pos}\leftrightarrow n_{neg}}$} & \multicolumn{3}{c}{15-1 (6 steps)}\\
    & & & old & new & all\\
    \hline
     & & & 79.5 & 49.8 & 72.4\\
     &\checkmark & & 79.6 & 50.6 & 72.7\\
     & &\checkmark & 79.6 & 50.3 & 72.6\\
     & \checkmark&\checkmark & 79.8 & 51.0 & 72.9\\
     \checkmark& \checkmark& & 79.8 & 51.4 & 73.0\\
     \checkmark&\checkmark &\checkmark & 79.9 & 51.9 & \textbf{73.2}\\
    \hline
  \end{tabular}
  \caption{
    Ablation study results for the component of CPD loss on PASCAL VOC \textit{overlapped} 15-1 (6 steps).
  }
  \label{tab4}  
\end{table}

\subsection{Ablation Study}
\textbf{Method components.} we present an evaluation of each component of the proposed Adapter in Tab.~\ref{tab3}. The "baseline" denotes the model using fixed prototype replay strategy and knowledge distillation in~\cite{star} with mBCE loss, which already achieves competitive performance. From the comparison between the first and second row in Tab.~\ref{tab3}, we can see that the ADC strategy improves the both performance of old and new classes (0.5 and 1.1 mIoU respectively), which arises from ADC compensating the representation deviation toward old-class and providing the appropriate features in new-class classifiers. Finally, our ADC strategy boosts the mIoU of all classes by 0.6, further demonstrating the feasibility of it. Comparing the second row with the third row, the effectiveness of UAC loss can be recognized. Specifically, UAC obtained gains of 1.3 in terms of new-class mIoU and 0.5 for all classes. This benefit by the constraint on uncertainty predictions, thereby compressing each class representation to a compact prototype and reserve latent space for learning new classes. Removing the CPD leads to the performance declining by 2.1 and 0.8 in terms of mIoU (row 3 vs. row 4) for new and all classes, respectively. Without the CPD loss, it is challenging to differentiate between representations of similar classes. This suggests that our CPD loss is more effective for generating distinguished features of each class and facilitating the effectiveness of the adaptive prototype replay strategy.

\textbf{Components of the CPD.} We present an ablation study of the CPD loss components on the PASCAL VOC \textit{overlapped} 15-1 scenario, as shown in Table~\ref{tab4}. The term "Copro" indicates that the old-class prototypes used in $\mathcal{L}_{n{\leftrightarrow}o}$ are compensated by our ADC strategy. The results in the table show that, with fixed old-class prototypes, the use of the two constraint terms results in a 0.5\% increase for all classes. When the old-class prototypes are updated using the ADC strategy, the mIoU improves by an additional 0.3\%. Each of the two loss terms also contributes to the enhanced performance.

\textbf{Generality of the UAC.} We applied our UAC loss to
two representative CISS methods, namely MIB~\cite{mib} and PLOP~\cite{plop}. Compared to MiB, our UAC improves performance by 5.3\%, with increases of 5.4\% and 3.9\% in mIoU for old and new classes on PASCAL VOC \textit{overlapped} 15-1 scenario, respectively. In additional, implementation of PLOP with our UAC also have 4.6\% performance improvement. These results provide convincing evidence that our insight of reducing prediction uncertainty is effective and generalizes well across different CISS methods.

\textbf{Ablation of Hyper-parameters.} Tab.~\ref{tab6} illustrates the influence of hyper-parameters: the weight of proposed loss terms $\beta$ and $\gamma$, and threshold $\tau$ in method. The results show that, in most cases, our approach is not highly sensitive to the choice of hyperparameters.

\begin{table}[t!]
  \centering
\footnotesize
  \begin{tabular}{c|ccc}
    \hline
      \multirow{2}*{Method} & \multicolumn{3}{c}{15-1 (6 steps)}\\
    & old & new & all\\
    \hline
      MiB & 35.1 & 13.5 & 29.7\\
      MiB+UAC & 40.5 & 17.4 & \textbf{35.0}\\
      \hline
      PLOP& 65.1 & 21.1 & 54.6\\
      PLOP+UAC& 66.1 & 30.0 & \textbf{59.2}\\
    \hline
    DKD&78.1	&42.7 &69.7\\
      DKD+UAC& 78.5&46.4&\textbf{70.9}\\
      \hline
        CoinSeg& 80.6 &36.2	&70.1\\
      CoinSeg+UAC& 80.8	&40.9&\textbf{71.3}\\
      \hline
  \end{tabular}
  \caption{
     UAC in other representative CISS methods. UAC
consistently enhances segmentation performance.}
  \label{tab5}  
\end{table}

\begin{table}[t!]
  \centering
  \footnotesize
  \begin{tabular}{cc|cc|cc}
    \hline
    \multirow{2}*{$\beta$} & 15-1 &\multirow{2}*{$\gamma$} & 15-1 & \multirow{2}*{$\tau$} & 15-1\\
    & (6 steps) & & (6 steps)& &(6 steps)\\
    \hline
     0.01 & 72.6 &0.01 & 72.5 & 0.5 & 72.3\\
     0.05  & 73.0 &0.05 &  \textbf{73.2} &0.6 &72.8\\
     0.1 & \textbf{73.2} &0.1 & 72.9 &0.7 &\textbf{73.2}\\
     0.5 & 72.9 &0.5 & 72.1 &0.8 &73.0\\
     1 & 72.7 &1 & 71.8 &0.9 &72.5\\
    \hline
  \end{tabular}
  \caption{
    Search of hyper-parameters: the weights of loss $\beta$ and $\gamma$, and threshold $\tau$. mIoU for all classes are reported}
  \label{tab6} 
  
\end{table}

\section{Conclusion}
\label{co}
In this paper, we propose a novel method, called Adapter, 
designed to address representation deviation in prototype replay based CISS methods. First, we introduce an adaptive deviation compensation strategy to estimate representation deviation between incremental models and update the stored prototype used for replay. Second, we employ an uncertainty-aware constraint loss and a compensation-based prototype discriminative loss to aggregate the representation of each class and enhance the discrimination of prototypes. Finally, experiments demonstrate the effectiveness of our proposed Adapter, which achieves remarkable performance, especially on challenging multi-step incremental settings, outperforming the previous state-of-the-art.

\section{Acknowledgments}
This work was supported by the Hubei Provincial Natural Science Foundation of China No.2022CFA055, the National Natural Science Foundation of China No.62176097.

\bibliography{aaai25}

\begin{thebibliography}{34}
\providecommand{\natexlab}[1]{#1}

\bibitem[{Baek et~al.(2022)Baek, Oh, Lee, Lee, and Ham}]{dkd}
Baek, D.; Oh, Y.; Lee, S.; Lee, J.; and Ham, B. 2022.
\newblock Decomposed knowledge distillation for class-incremental semantic
  segmentation.
\newblock \emph{Advances in Neural Information Processing Systems}, 35.

\bibitem[{Cermelli et~al.(2020)Cermelli, Mancini, Bulo, Ricci, and
  Caputo}]{mib}
Cermelli, F.; Mancini, M.; Bulo, S.~R.; Ricci, E.; and Caputo, B. 2020.
\newblock Modeling the background for incremental learning in semantic
  segmentation.
\newblock In \emph{Proceedings of the IEEE/CVF Conference on Computer Vision
  and Pattern Recognition}, 9233--9242.

\bibitem[{Cha et~al.(2021)Cha, Yoo, Moon et~al.}]{ssul}
Cha, S.; Yoo, Y.; Moon, T.; et~al. 2021.
\newblock Ssul: Semantic segmentation with unknown label for exemplar-based
  class-incremental learning.
\newblock \emph{Advances in neural information processing systems}, 34.

\bibitem[{Chen et~al.(2023)Chen, Cong, Luo, Ip, and Kwong}]{star}
Chen, J.; Cong, R.; Luo, Y.; Ip, H.; and Kwong, S. 2023.
\newblock Saving 100x Storage: Prototype Replay for Reconstructing Training
  Sample Distribution in Class-Incremental Semantic Segmentation.
\newblock \emph{Advances in Neural Information Processing Systems}, 36.

\bibitem[{Chen et~al.(2017)Chen, Papandreou, Kokkinos, Murphy, and
  Yuille}]{deeplab}
Chen, L.-C.; Papandreou, G.; Kokkinos, I.; Murphy, K.; and Yuille, A.~L. 2017.
\newblock Deeplab: Semantic image segmentation with deep convolutional nets,
  atrous convolution, and fully connected crfs.
\newblock \emph{IEEE Transactions on Pattern Analysis and Machine
  Intelligence}, 40(4): 834--848.

\bibitem[{Douillard et~al.(2021)Douillard, Chen, Dapogny, and Cord}]{plop}
Douillard, A.; Chen, Y.; Dapogny, A.; and Cord, M. 2021.
\newblock Plop: Learning without forgetting for continual semantic
  segmentation.
\newblock In \emph{Proceedings of the IEEE/CVF conference on computer vision
  and pattern recognition}, 4040--4050.

\bibitem[{Douillard et~al.(2020)Douillard, Cord, Ollion, Robert, and
  Valle}]{podnet}
Douillard, A.; Cord, M.; Ollion, C.; Robert, T.; and Valle, E. 2020.
\newblock Podnet: Pooled outputs distillation for small-tasks incremental
  learning.
\newblock In \emph{Proceedings of the European conference on computer vision
  (ECCV)}, 86--102. Springer.

\bibitem[{Everingham et~al.(2010)Everingham, Van~Gool, Williams, Winn, and
  Zisserman}]{pascal}
Everingham, M.; Van~Gool, L.; Williams, C.~K.; Winn, J.; and Zisserman, A.
  2010.
\newblock The pascal visual object classes (voc) challenge.
\newblock \emph{International journal of computer vision}, 88: 303--338.

\bibitem[{Gou et~al.(2021)Gou, Yu, Maybank, and Tao}]{kd}
Gou, J.; Yu, B.; Maybank, S.~J.; and Tao, D. 2021.
\newblock Knowledge distillation: A survey.
\newblock \emph{International Journal of Computer Vision}, 129(6): 1789--1819.

\bibitem[{He et~al.(2016)He, Zhang, Ren, and Sun}]{deep}
He, K.; Zhang, X.; Ren, S.; and Sun, J. 2016.
\newblock Deep residual learning for image recognition.
\newblock In \emph{Proceedings of the IEEE conference on computer vision and
  pattern recognition}, 770--778.

\bibitem[{Jung et~al.(2020)Jung, Ahn, Cha, and Moon}]{AGS}
Jung, S.; Ahn, H.; Cha, S.; and Moon, T. 2020.
\newblock Continual learning with node-importance based adaptive group sparse
  regularization.
\newblock \emph{Advances in neural information processing systems}, 33.

\bibitem[{Kirkpatrick et~al.(2017)Kirkpatrick, Pascanu, Rabinowitz, Veness,
  Desjardins, Rusu, Milan, Quan, Ramalho, Grabska-Barwinska et~al.}]{ewc}
Kirkpatrick, J.; Pascanu, R.; Rabinowitz, N.; Veness, J.; Desjardins, G.; Rusu,
  A.~A.; Milan, K.; Quan, J.; Ramalho, T.; Grabska-Barwinska, A.; et~al. 2017.
\newblock Overcoming catastrophic forgetting in neural networks.
\newblock volume 114, 3521--3526. National Acad Sciences.

\bibitem[{Li and Hoiem(2017)}]{lwf}
Li, Z.; and Hoiem, D. 2017.
\newblock Learning without forgetting.
\newblock \emph{IEEE transactions on pattern analysis and machine
  intelligence}, 40(12): 2935--2947.

\bibitem[{Lopez-Paz and Ranzato(2017)}]{exam}
Lopez-Paz, D.; and Ranzato, M. 2017.
\newblock Gradient episodic memory for continual learning.
\newblock \emph{Advances in neural information processing systems}, 30.

\bibitem[{Maracani et~al.(2021)Maracani, Michieli, Toldo, and
  Zanuttigh}]{recall}
Maracani, A.; Michieli, U.; Toldo, M.; and Zanuttigh, P. 2021.
\newblock Recall: Replay-based continual learning in semantic segmentation.
\newblock In \emph{Proceedings of the IEEE/CVF international conference on
  computer vision}, 7026--7035.

\bibitem[{Masana et~al.(2022)Masana, Liu, Twardowski, Menta, Bagdanov, and Van
  De~Weijer}]{increlearning}
Masana, M.; Liu, X.; Twardowski, B.; Menta, M.; Bagdanov, A.~D.; and Van
  De~Weijer, J. 2022.
\newblock Class-incremental learning: survey and performance evaluation on
  image classification.
\newblock \emph{IEEE Transactions on Pattern Analysis and Machine
  Intelligence}, 45(5): 5513--5533.

\bibitem[{McCloskey and Cohen(1989)}]{catastrophic}
McCloskey, M.; and Cohen, N.~J. 1989.
\newblock Catastrophic interference in connectionist networks: The sequential
  learning problem.
\newblock In \emph{Psychology of learning and motivation}, volume~24, 109--165.
  Elsevier.

\bibitem[{Michieli and Zanuttigh(2019)}]{ilt}
Michieli, U.; and Zanuttigh, P. 2019.
\newblock Incremental learning techniques for semantic segmentation.
\newblock In \emph{Proceedings of the IEEE/CVF international conference on
  computer vision workshops}.

\bibitem[{Michieli and Zanuttigh(2021)}]{sdr}
Michieli, U.; and Zanuttigh, P. 2021.
\newblock Continual semantic segmentation via repulsion-attraction of sparse
  and disentangled latent representations.
\newblock In \emph{Proceedings of the IEEE/CVF conference on computer vision
  and pattern recognition}, 1114--1124.

\bibitem[{Oh, Baek, and Ham(2022)}]{alife}
Oh, Y.; Baek, D.; and Ham, B. 2022.
\newblock Alife: Adaptive logit regularizer and feature replay for incremental
  semantic segmentation.
\newblock \emph{Advances in Neural Information Processing Systems}, 35.

\bibitem[{Phan et~al.(2022)Phan, Phung, Tran-Thanh, Bouzerdoum
  et~al.}]{reminder}
Phan, M.~H.; Phung, S.~L.; Tran-Thanh, L.; Bouzerdoum, A.; et~al. 2022.
\newblock Class similarity weighted knowledge distillation for continual
  semantic segmentation.
\newblock In \emph{Proceedings of the IEEE/CVF Conference on Computer Vision
  and Pattern Recognition}, 16866--16875.

\bibitem[{Rebuffi et~al.(2017)Rebuffi, Kolesnikov, Sperl, and Lampert}]{icarl}
Rebuffi, S.-A.; Kolesnikov, A.; Sperl, G.; and Lampert, C.~H. 2017.
\newblock icarl: Incremental classifier and representation learning.
\newblock In \emph{Proceedings of the IEEE conference on Computer Vision and
  Pattern Recognition}, 2001--2010.

\bibitem[{Shin et~al.(2017)Shin, Lee, Kim, and Kim}]{replay}
Shin, H.; Lee, J.~K.; Kim, J.; and Kim, J. 2017.
\newblock Continual learning with deep generative replay.
\newblock \emph{Advances in neural information processing systems}, 30.

\bibitem[{Shmelkov, Schmid, and Alahari(2017)}]{indetector}
Shmelkov, K.; Schmid, C.; and Alahari, K. 2017.
\newblock Incremental learning of object detectors without catastrophic
  forgetting.
\newblock In \emph{Proceedings of the IEEE international conference on computer
  vision}, 3400--3409.

\bibitem[{Yang et~al.(2022)Yang, Fini, Xu, Rota, Ding, Nabi, Alameda-Pineda,
  and Ricci}]{ucd}
Yang, G.; Fini, E.; Xu, D.; Rota, P.; Ding, M.; Nabi, M.; Alameda-Pineda, X.;
  and Ricci, E. 2022.
\newblock Uncertainty-aware contrastive distillation for incremental semantic
  segmentation.
\newblock \emph{IEEE Transactions on Pattern Analysis and Machine
  Intelligence}, 45(2): 2567--2581.

\bibitem[{Yoon et~al.(2018)Yoon, Yang, Lee, and Hwang}]{den}
Yoon, J.; Yang, E.; Lee, J.; and Hwang, S.~J. 2018.
\newblock Lifelong Learning with Dynamically Expandable Networks.
\newblock In \emph{International Conference on Learning Representations}.

\bibitem[{Yu et~al.(2020)Yu, Twardowski, Liu, Herranz, Wang, Cheng, Jui, and
  Weijer}]{sdc}
Yu, L.; Twardowski, B.; Liu, X.; Herranz, L.; Wang, K.; Cheng, Y.; Jui, S.; and
  Weijer, J. v.~d. 2020.
\newblock Semantic drift compensation for class-incremental learning.
\newblock In \emph{Proceedings of the IEEE/CVF conference on computer vision
  and pattern recognition}, 6982--6991.

\bibitem[{Zenke, Poole, and Ganguli(2017)}]{si}
Zenke, F.; Poole, B.; and Ganguli, S. 2017.
\newblock Continual learning through synaptic intelligence.
\newblock In \emph{International conference on machine learning}, 3987--3995.
  PMLR.

\bibitem[{Zhang et~al.(2022{\natexlab{a}})Zhang, Xiao, Liu, Chen, and
  Cheng}]{rcil}
Zhang, C.-B.; Xiao, J.-W.; Liu, X.; Chen, Y.-C.; and Cheng, M.-M.
  2022{\natexlab{a}}.
\newblock Representation compensation networks for continual semantic
  segmentation.
\newblock In \emph{Proceedings of the IEEE/CVF Conference on Computer Vision
  and Pattern Recognition}, 7053--7064.

\bibitem[{Zhang et~al.(2022{\natexlab{b}})Zhang, Gao, Fang, Jiao, and
  Wei}]{microseg}
Zhang, Z.; Gao, G.; Fang, Z.; Jiao, J.; and Wei, Y. 2022{\natexlab{b}}.
\newblock Mining unseen classes via regional objectness: A simple baseline for
  incremental segmentation.
\newblock \emph{Advances in neural information processing systems}, 35:
  24340--24353.

\bibitem[{Zhang et~al.(2023)Zhang, Gao, Jiao, Liu, and Wei}]{coinseg}
Zhang, Z.; Gao, G.; Jiao, J.; Liu, C.~H.; and Wei, Y. 2023.
\newblock Coinseg: Contrast inter-and intra-class representations for
  incremental segmentation.
\newblock In \emph{Proceedings of the IEEE/CVF International Conference on
  Computer Vision}, 843--853.

\bibitem[{Zhao, Yuan, and Shi(2023)}]{Idec}
Zhao, D.; Yuan, B.; and Shi, Z. 2023.
\newblock Inherit With Distillation and Evolve With Contrast: Exploring Class
  Incremental Semantic Segmentation Without Exemplar Memory.
\newblock \emph{IEEE Transactions on Pattern Analysis and Machine
  Intelligence}, 45(10): 11932--11947.

\bibitem[{Zhao et~al.(2021)Zhao, Wang, Fu, Wu, and Li}]{meic}
Zhao, H.; Wang, H.; Fu, Y.; Wu, F.; and Li, X. 2021.
\newblock Memory-efficient class-incremental learning for image classification.
\newblock \emph{IEEE Transactions on Neural Networks and Learning Systems},
  33(10): 5966--5977.

\bibitem[{Zhou et~al.(2017)Zhou, Zhao, Puig, Fidler, Barriuso, and
  Torralba}]{Ade}
Zhou, B.; Zhao, H.; Puig, X.; Fidler, S.; Barriuso, A.; and Torralba, A. 2017.
\newblock Scene parsing through ade20k dataset.
\newblock In \emph{Proceedings of the IEEE conference on computer vision and
  pattern recognition}, 633--641.

\end{thebibliography}

\section{Appendix}
 In this supplementary section, we present additional information on Adapter, encompassing more details on expanded experimental results. The supplementary material is organized as follows:
 \begin{enumerate}
     \item Section A introduces quantitative results of Adapter, including short overlapped settings, more ablation studies, effect of example-replay, and comparison on \textit{disjoint} settings.
     \item Section B introduces qualitative results of Adapter, referring the visualization of each steps on VOC 15-1 and ADE 100-10.
 \end{enumerate}

\section{A. More Quantitative Results}
\subsection{Results on the Short Overlapped Settings}

In Tab.~\ref{tab7}, we present an additional quantitative comparison between our Adapter and previous state-of-the-art methods (MiB~\cite{mib}, PLOP~\cite{plop}, SSUL~\cite{ssul}, DKD~\cite{dkd}, UCD~\cite{ucd}, REMIND~\cite{reminder}, MicroSeg~\cite{microseg}, ALIFE~\cite{alife}, STAR~\cite{star}) under the short \textit{overlapped} settings. Our method achieves a remarkable balance between stability and scalability, with mIoU scores of 78.0 for old classes and 50.7 for new classes. It also demonstrates a 0.2 performance improvement over the closest competitor~\cite{star}. Additionally, our model shows further improvement, outperforming the state-of-the-art by 0.4 in the 15-5 setting with respect to the all classes mIoU. These results in the table further prove that our method can also improve the performance of short \textit{overlapped} scenarios.
\begin{table}[htbp]
  \centering
  \footnotesize
  \begin{tabular}{c|ccc|ccc}
    \hline
    \multirow{2}*{Method} 
     & \multicolumn{3}{c|}{19-1 (2 steps)} &  \multicolumn{3}{c}{15-5 (2 steps)} \\ 
    & old & new & all & old & new & all \\
    \hline

    MiB & 70.2 & 22.1 & 67.8 & 75.5 & 49.4 & 69.0 \\

    PLOP & 75.4 & 37.4 & 73.5 &  75.7 & 51.7 & 70.1 \\

    SSUL & 77.7 & 29.7 & 75.4 & 77.8 & 50.1 & 71.2 \\

    DKD & 77.8 & 41.5 & 76.0 & 78.8 & 58.2 & 73.9 \\

    UCD & 75.9 & 39.5 & 74.0 & 75.0 & 51.8 & 69.2 \\

    REMIND & 76.5 & 32.3 & 74.4 & 76.1 & 50.7 & 70.1 \\

    MicroSeg & \textbf{78.8} & 14.0 & 75.7 & \textbf{80.4} & 52.8 & 73.8 \\

    ALIFE & 76.6 & \underline{49.3} & 75.3 & 77.1 & 52.5 &  71.3 \\

    STAR & 78.0 & 47.1 & \underline{76.5} & 79.5 & \underline{58.9} & \underline{74.6} \\

    Ours & \underline{78.0} & \textbf{50.7} & \textbf{76.7} & \underline{79.7} & \textbf{59.7} & \textbf{75.0} \\
    \hline
  \end{tabular}
  \caption{
    Quantitative results on the validation set of PASCAL VOC~\cite{pascal} for the short \textit{overlapped} settings (i.e, 19-1 and 15-5).
  }
  \label{tab7}
\end{table}

\begin{table}[htbp]
  \centering
  \footnotesize
  \begin{tabular}{c|c|c|c|ccc}
    \hline
    \multirow{2}*{Baseline} & \multirow{2}*{ADC}& \multirow{2}*{UAC}& \multirow{2}*{CPD}&\multicolumn{3}{c}{15-1 (6 steps)}\\
    & & & & old & new & all\\
    \hline
    \checkmark & & & & 78.7 & 47.4 & 71.3\\
    \checkmark & & \checkmark& & 78.9&49.1&71.8 \\
    \checkmark &  &  &  \checkmark& 79.6 & 49.3 & 72.4\\
    \checkmark & & \checkmark & \checkmark& 79.7 & 50.3 & 72.7\\
    \checkmark & \checkmark & & \checkmark& 79.8 & 50.8 & 72.9\\
    \hline
  \end{tabular}
  \caption{
    More ablation study of our method components on PASCAL VOC \textit{overlapped} 15-1 (6 steps).
  }
  \label{tab8}  
\end{table}

\begin{table*}[t!]
  \centering
  \begin{tabular}{c|ccc|ccc|ccc}
    \hline
    \multirow{2.5}*{Method} 
     & \multicolumn{3}{c|}{19-1 (2 steps)} &  \multicolumn{3}{c|}{15-5 (2 steps)}  &  \multicolumn{3}{c}{15-1 (6 steps)} \\ 
    & old & new & all & old & new & all & old & new & all \\
    \hline

    MiB &  69.6 & 25.6 & 67.4 & 71.8 & 43.3 & 64.7 & 46.2 & 12.9 & 37.9 \\

    PLOP & 75.4 & 38.9 & 73.6 &  71.0 & 42.8 & 64.3 & 57.7 & 13.7 & 46.5\\

    SDR &69.9 &37.3 &68.4 &73.5 &47.3 &67.2 &  59.2 &12.9 &48.1 \\
    
    SSUL & 77.4 & 22.4 & 74.8 & 76.4 & 45.6 & 69.1 & 74.0 & 32.2 & 64.0\\

    DKD & 77.4 & 43.6 & 75.8 & 77.6 & 54.1 & 72.0 & 76.3 & 39.4 & 67.5\\

    UCD & 73.4 & 33.7 & 71.5 & 71.9 & 49.5 & 66.2 & 53.1 & 13.0 & 42.9\\

    RCIL & - & - & - &  75.0 & 42.8 & 67.3 &  66.1 & 18.2 & 54.7 \\

    STAR & \underline{77.9} & \underline{44.7} &\underline{76.3}  & \underline{78.3} & \underline{57.7} & \underline{73.4} & \underline{78.2} & \underline{47.6} & \underline{70.9} \\

    Ours & \textbf{78.0} & \textbf{46.1} & \textbf{76.5} &\textbf{78.9} &\textbf{58.2} &\textbf{73.9} &  \textbf{78.6} 	& \textbf{49.0} & \textbf{71.5} \\
    \hline
  \end{tabular}
 \caption{Quantitative results on the validation set of PASCAL VOC for the \textit{disjoint} settings.}
  \label{tab10} 
\end{table*}

\subsection{More Ablation Studies}
Adapter involves multiple loss functions, such as uncertainty-aware constraint (UAC) loss and compensation-based prototype similarity discriminative (CPD) loss. We further explore the effects of them on Adapter in the Table \ref{tab8}. Form the results in the table, with uncertainty-aware constraint loss (row 2), the baseline obtain the 0.5 improvement in terms of MIoU for all classes. In a combination of these two losses, Adapter achieve 72.7 in terms of MIoU. Results show that UAC and CPD again performance improvement compared to baseline, verifying the effectiveness of two proposed loss.

\subsection{Effects of Examples Memory}
In the CISS community, some works focused on the replay-based method that replay the past knowledge, like STAR~\cite{star}. There are also some works focused on memory-based method that directly store past data, like SSUL~\cite{ssul}, MicroSeg~\cite{microseg}. Objectively, there is a gap between the generated distribution and the real data. Thus, STAR extended the memory-based version called STAR-M with storing tiny past samples. Following the STAR, we also compare the extended version Adapter-M with other methods on VOC 15-1. As shown in Table~\ref{tab9},
 our Adapter (the 5th row) achieves better performance on VOC 15-1, even when compared with
 previous work using the sampling strategy. When equipped with the memory sampling method, i.e., Adapter-M,it undoubtedly achieves state-of-the-art performance. The Adapter-M achieved performance gains of 0.4 (74.8 vs. 74.4) even if it only stored half the memory. Specially, MicroSeg used additional proposal maps to help model train more accurately.

\begin{table}[htbp]
  \centering
\footnotesize
  \begin{tabular}{c|ccc}
    \hline
      \multirow{2}*{Method} & \multicolumn{3}{c}{15-1 (6 steps)}\\
    & old & new & all\\
    \hline
      SSUL & 77.3 &36.6 &67.6\\
      SSUL-M(100) &78.4&49.0&71.4\\
      \hline
      MicroSeg& 80.1 &36.8 &69.8 \\
      MicroSeg-M(100)& 81.3&52.5&74.4\\
    \hline
    STAR&79.4 &50.3 &72.5\\
      STAR-M(50)& 79.5&55.6&73.8\\
      \hline
        Ours& 79.9 &51.9 &73.2\\
      Ours-M(50)& 80.6&56.3&74.8\\
      \hline
  \end{tabular}
  \caption{Comparisons of Adapter using example memory}
  \label{tab9}  
\end{table}

 \subsection{Impact of Transformer Backbone}
Although the architecture that model uses is not our focus, we integrate our method to CoinSeg that based on Swin-B for VOC 15-1. The results in the Table~ \ref{tab11} further verify that our method is effective for transformer-based models.

\begin{table}[htbp]
  \centering
\footnotesize
  \begin{tabular}{c|c|ccc}
    \hline
      \multirow{2}*{Method} & \multirow{2}*{Backbone}& \multicolumn{3}{c}{15-1 (6 steps)}\\
    & & old & new & all\\
    \hline
      CoinSeg& Swin-B & 82.7&52.5&75.5\\
      Ours&Swin-B &	83.3	&60.1	&77.8\\
      \hline
      
  \end{tabular}
  \caption{Adapter with Swin-B as backbone}
  \label{tab11}  
\end{table}

\subsection{Comparison Under the Disjoint Settings}
Additionally, we also report the results compare to previous methods (MiB~\cite{mib}, PLOP~\cite{plop}, SSUL~\cite{ssul}, DKD~\cite{dkd}, UCD~\cite{ucd}, SDR~\cite{sdr}, RCIL~\cite{rcil}, STAR~\cite{star}) on the three Pascal VOC \textit{disjoint} settings (i.e., 19-1, 15-5, 15-1) in Tab.~\ref{tab10}. In contrast to the \textit{overlapped} setting, the training dataset during incremental steps under the \textit{disjoint} setting includes only the samples from the current and previous steps, whereas the \textit{overlapped} setting may encompass future classes. From the results in the table, we can see that our method obtains a similar performance improvement as in the overlapped settings. Specifically, our method outperforms the state-of-the-art by
0.2\% and 0.5\% on the short 19-1 and 15-5 settings with 2 incremental steps, respectively. In the 15-1 setting with 6 steps, we achieved a 0.4\% and 1.4\% increase in the "old" and "new" mIoU, respectively, resulting in a final mIoU of 71.5\%. This further demonstrates the effectiveness and robustness of our method under different incremental scenarios.

\section{B. More Qualitative Results}

\begin{figure*}[t]
\centering
\includegraphics[width =0.8\textwidth]{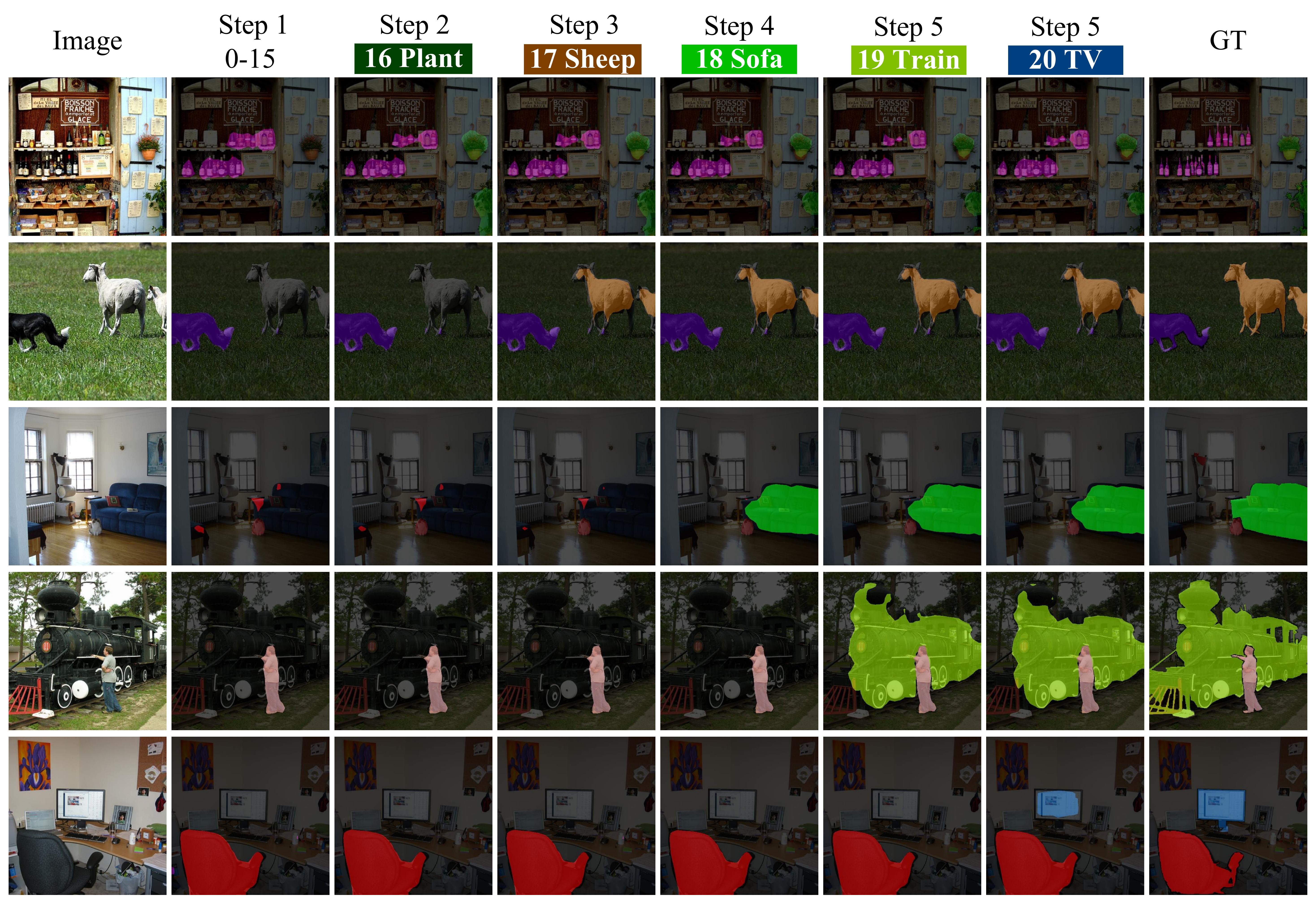}
\caption{Qualitative results for the overlapped 15-1 setting on Pascal VOC 2012. \colorbox{plant}{\textcolor{white}{\textit{Plant}}}, \colorbox{sheep}{\textcolor{white}{\textit{Sheep}}}, \colorbox{sofa}{\textcolor{white}{\textit{Sofa}}}, \colorbox{train}{\textcolor{white}{\textit{Train}}}, and \colorbox{tv}{\textcolor{white}{\textit{TV}}} are new coming classes in corresponding steps.}
\label{fig4}
\end{figure*}

\begin{figure*}[t]
\centering
\includegraphics[width =0.8\textwidth]{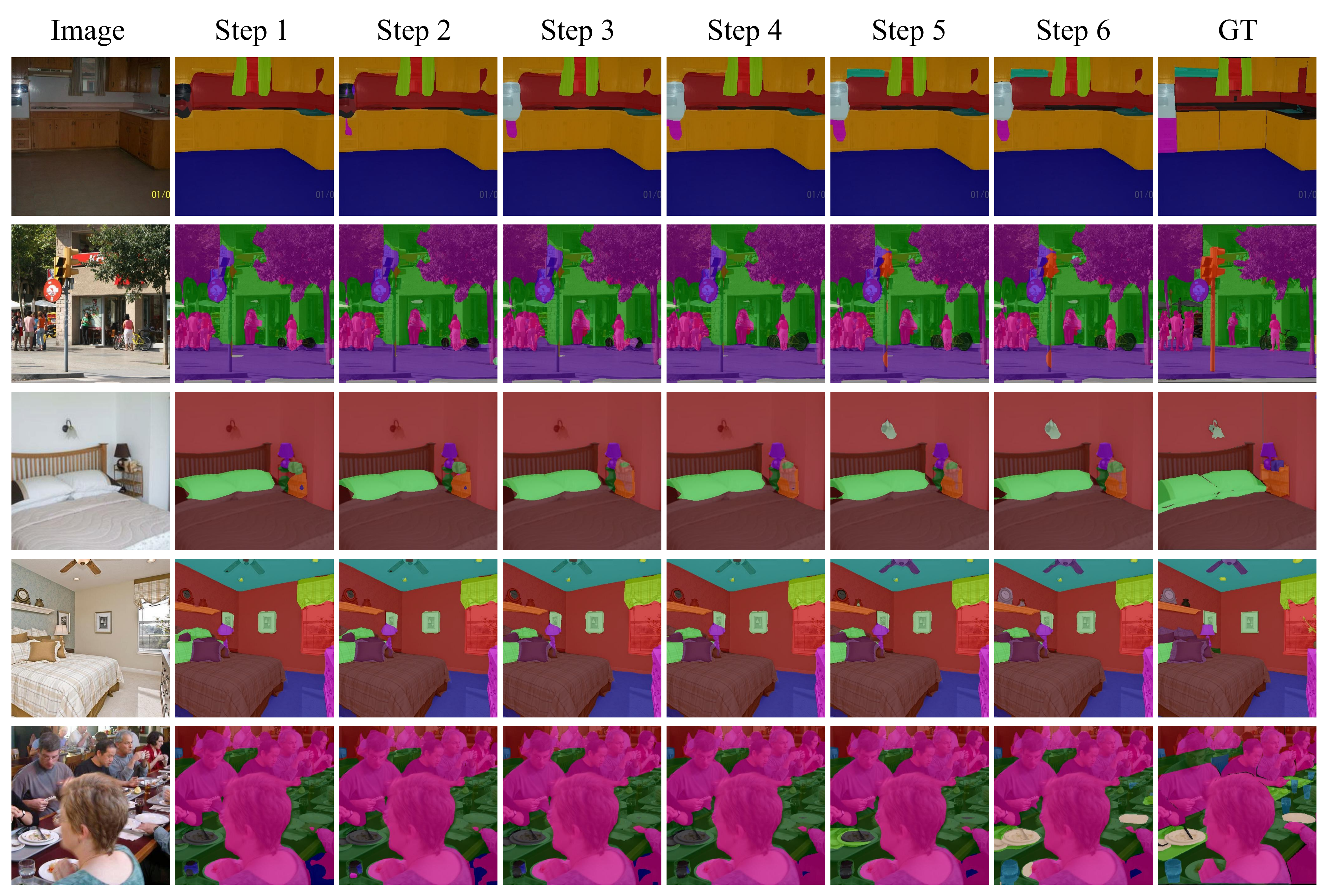}
\caption{ Qualitative results for the overlapped 100-10 setting on ADE20K. \colorbox{hood}{\textcolor{white}{\textit{Hood}}}, \colorbox{oven}{\textcolor{white}{\textit{Oven}}}, 
\colorbox{traffic}{\textcolor{white}{\textit{Traffic light}}}, \colorbox{sconce}{\textcolor{white}{\textit{Sconce}}}, \colorbox{fan}{\textcolor{white}{\textit{Fan}}}, \colorbox{plate}{\textcolor{white}{\textit{Plate}}}, and \colorbox{glass}{\textcolor{white}{\textit{Glass}}} are new coming classes in corresponding steps.}
\label{fig5}
\end{figure*}

We visualize the results of our method on different steps of PASCAL VOC~\cite{pascal} \textit{overlapped} 15-1 (6 steps) and ADE20K~\cite{Ade} \textit{overlapped} 100-10 (6 steps) to unveil more details. For Pascal VOC, as shown in Fig.~\ref{fig4}, our method effectively predicts five new classes (e.g., sofa, train, TV) while minimizing forgetting of previously learned classes. Notably, in the third row of Fig.~\ref{fig5}, some pixels initially misclassified as chairs in earlier steps were correctly identified as sofas in subsequent steps. Additionally, our method successfully maintain the correct predicts occur in the previous steps (like \textit{dog}, \textit{person}, and \textit{chair} in the 
row 2, 4, 5 respectively). This demonstrates the effectiveness of our adaptive representation deviation compensation strategy and two constraint losses. The ADE20K dataset, with its extensive number of classes to be learned at both the initial and incremental steps, and the relatively small proportion of some classes, still allows us to successfully differentiate new classes (as shown in Fig.~\ref{fig5}) without compromising the performance on old classes. For example, in the row 2, although the \textit{traffic light} is very narrow and occupies a minimal area, our method effectively identifies most of its region successfully. Moreover, the pixels of new classes like \textit{sconce}, \textit{fan}, and \textit{plate} in the correspond rows also are correctly recognized. 
\end{document}